\documentclass[10pt,twocolumn,letterpaper]{article}

\usepackage{cvpr}
\usepackage{times}
\usepackage{epsfig}
\usepackage{graphicx}
\usepackage{amsmath}
\usepackage{amssymb}

\usepackage{dsfont}
\usepackage{fancyhdr}
\graphicspath{{./figure/}}

\usepackage[pagebackref=true,breaklinks=true,letterpaper=true,colorlinks,bookmarks=false]{hyperref}

\cvprfinalcopy % *** Uncomment this line for the final submission

\def\cvprPaperID{119} % *** Enter the CVPR Paper ID here

\begin{document}

%%%%%%%%% TITLE
\title{Person Re-identification by Local Maximal Occurrence\\Representation and Metric Learning}

\author{Shengcai Liao, Yang Hu, Xiangyu Zhu, and Stan Z. Li\\
{\small Center for Biometrics and Security Research, National Laboratory of Pattern Recognition}\\
{\small Institute of Automation, Chinese Academy of Sciences, 95 Zhongguancun East Road, Beijing 100190, China}\\
{\tt\small \{scliao,yhu,xiangyu.zhu,szli\}@nlpr.ia.ac.cn}
}

\maketitle
\thispagestyle{fancy}\chead{\small To appear in the IEEE International Conference on Computer Vision and Pattern Recognition (CVPR 2015), Boston, USA, June 2015}

%%%%%%%%% ABSTRACT
\begin{abstract}
Person re-identification is an important technique towards automatic search of a person's presence in a surveillance video. Two fundamental problems are critical for person re-identification, feature representation and metric learning. An effective feature representation should be robust to illumination and viewpoint changes, and a discriminant metric should be learned to match various person images. In this paper, we propose an effective feature representation called Local Maximal Occurrence (LOMO), and a subspace and metric learning method called Cross-view Quadratic Discriminant Analysis (XQDA). The LOMO feature analyzes the horizontal occurrence of local features, and maximizes the occurrence to make a stable representation against viewpoint changes. Besides, to handle illumination variations, we apply the Retinex transform and a scale invariant texture operator. To learn a discriminant metric, we propose to learn a discriminant low dimensional subspace by cross-view quadratic discriminant analysis, and simultaneously, a QDA metric is learned on the derived subspace. We also present a practical computation method for XQDA, as well as its regularization. Experiments on four challenging person re-identification databases, VIPeR, QMUL GRID, CUHK Campus, and CUHK03, show that the proposed method improves the state-of-the-art rank-1 identification rates by 2.2\%, 4.88\%, 28.91\%, and 31.55\% on the four databases, respectively.
\end{abstract}

%%%%%%%%% BODY TEXT
\section{Introduction}
Person re-identification is a problem of finding a person from a gallery who has the same identity to the probe. This is a challenging problem because of big intra-class variations in illumination, pose or viewpoint, and occlusion. Many approaches have been proposed for person re-identification \cite{Vezzani-CSUR-2013,gong2014person}, which greatly advance this field.

Two fundamental problems are critical for person re-identification, feature representation and metric learning. An effective feature representation should be robust to illumination and viewpoint changes, and a discriminant metric should be learned to match various person images. Many efforts have been made along the two directions to tackle the challenge of person re-identification. For feature representation, several effective approaches have been proposed, for example, the ensemble of local features (ELF) \cite{Gray-ECCV-08}, SDALF \cite{Bazzani-CVIU-2013-SDALF}, kBiCov \cite{Ma-IVC-2014-kBiCov}, fisher vectors (LDFV) \cite{ma2012local}, salience match \cite{zhao2013person}, and mid-level filter \cite{Zhao-CVPR-2014-MidLF}. These hand-crafted or learning based descriptors have made impressive improvements over robust feature representation, and advanced the person re-identification research. However, how to design or learn a robust feature for the person re-identification challenge still remains an open problem.

Another aspect of person re-identification is how to learn a robust distance or similarity function to deal with the complex matching problem. Many metric learning algorithms have been proposed considering this aspect \cite{dikmen2011pedestrian,zheng2011person,kostinger2012large,Hirzer-ECCV-2012-RPLM,li2013learning}. In practice, many previous metric learning methods \cite{weinberger2006distance,davis2007information,dikmen2011pedestrian,Hirzer-ECCV-2012-RPLM,kostinger2012large,Pedagadi-CVPR-2013-LF} show a two-stage processing for metric learning, that is, the Principle Component Analysis (PCA) is first applied for dimension reduction, then metric learning is performed on the PCA subspace. %Zheng et al. \cite{zheng2011person} found that, for some metric learning algorithms like LMNN \cite{weinberger2006distance} and ITML \cite{davis2007information}, dimension reduction by PCA must be performed, otherwise they become intractable given the high dimensional feature space.
However, this two-stage processing may not be optimal for metric learning in low dimensional space, because samples of different classes may already be cluttered after the first stage.

In this paper, we propose an efficient feature representation called Local Maximal Occurrence (LOMO), and a subspace and metric learning method called Cross-view Quadratic Discriminant Analysis (XQDA). The LOMO feature analyzes the horizontal occurrence of local features, and maximizes the occurrence to make a stable representation against viewpoint changes. Besides, we find that applying the Retinex transform is useful to handle illumination variations in person re-identification. To learn a discriminant metric, we propose to learn a discriminant low dimensional subspace by cross-view quadratic discriminant analysis, and simultaneously, a QDA metric is learned on the derived subspace. We show that the problem can be formulated as a Generalized Rayleigh Quotient, and a closed-form solution can be obtained by the generalized eigenvalue decomposition. We also present a practical computation method for XQDA, as well as its regularization and dimension selection. The proposed method is shown to be effective and efficient through person re-identification experiments on four public databases, and we also demonstrate how the proposed components lead to improvements.

\section{Related Work}\label{sec:review}
Many existing person re-identification approaches try to build a robust feature representation which is both distinctive and robust for describing a person's appearance under various conditions \cite{Gray-ECCV-08,huexploring,DBLP:conf/cvpr/GheissariSH06,DBLP:conf/icdsc/HamdounMSS08,DBLP:conf/iccv/WangDSRT07,cheng2011custom}. Gray and Tao \cite{Gray-ECCV-08} proposed to use AdaBoost to select good features out of a set of color and texture features. Farenzena et al. \cite{DBLP:conf/cvpr/FarenzenaBPMC10} proposed the Symmetry-Driven Accumulation of Local Features (SDALF) method, where the symmetry and asymmetry property is considered to handle viewpoint variations. Ma et al. \cite{ma2012local} turned local descriptors into the Fisher Vector to produce a global representation of an image. Cheng et al. \cite{cheng2011custom} utilized the Pictorial Structures where part-based color information and color displacement were considered for person re-identification.
Recently, saliency information has been investigated for person re-identification \cite{zhao2013unsupervised,zhao2013person,liu2012person}, leading to a novel feature representation. In \cite{Wang-Regionlets-ICCV-2013}, a method called regionlets is proposed, which  picks a maximum bin from three random regions for object detection under deformation. In contrast, we propose to maximize the occurrence of each local pattern among all horizontal sub-windows to tackle viewpoint changes.

Besides robust features, metric learning has been widely applied for person re-identification \cite{weinberger2006distance,davis2007information,Guillaumin-ICCV-09,dikmen2011pedestrian,zheng2011person,kostinger2012large,Hirzer-ECCV-2012-RPLM,li2013learning}. Zheng et al. \cite{zheng2011person} proposed the PRDC algorithm, which optimizes the relative distance comparison. Hirzer et al. \cite{Hirzer-ECCV-2012-RPLM} proposed to relax the PSD constraint required in Mahalanobis metric learning, and obtained a simplified formulation that still showed promising performance. Li at al.\cite{li2013learning} proposed the learning of Locally-Adaptive Decision Functions (LADF) for person verification, which can be viewed as a joint model of a distance metric and a locally adapted thresholding rule. Prosser et al.\cite{prosser2010person} formulated the person re-identification problem as a ranking problem, and applied the RankSVM to learn a subspace. In \cite{Li-CVPR-2013-CUHK02}, local experts were considered to learn a common feature space for person re-identification across views.

Except a novel feature representation, the proposed XQDA algorithm is mostly related to Bayesian face \cite{Moghaddam-PR-2000}, KISSME \cite{kostinger2012large}, Linear Discriminant Analysis (LDA) \cite{Hastie-Book-2009}, local fisher discriminant analysis (LF) \cite{Pedagadi-CVPR-2013-LF}, and CFML \cite{Alipanahi-AAAI-2008}. XQDA can be seen as an extension of Bayesian face and KISSME, in that a discriminant subspace is further learned together with a metric. The LF method applies FDA together with PCA and LPP to derive a low dimensional yet discriminant subspace. The CFML algorithm aims at a different problem though learns a similar subspace to XQDA. However, both LF and CFML use the Euclidean distance on the derived subspace, while the proposed method considers a discriminant subspace as well as an integrated metric. For the traditional LDA method, though XQDA shares a similar generalized Rayleigh quotient formulation, they are essentially not equivalent, which is explained in \cite{Alipanahi-AAAI-2008}.

\section{Local Maximal Occurrence Feature}\label{sec:feature}
%Person re-identification is a challenging problem due to big intra-class variations in illumination, viewpoint, and occlusion. Therefore, robust feature representation is important for the description of person images \cite{Liu-ECCVW-2012}. We propose a feature representation called Local Maximal Occurrence (LOMO) for person re-identification, which addresses the illumination and viewpoint changes, as described below.

\subsection{Dealing with Illumination Variations}
\begin{figure}
\centering
\includegraphics[width=6mm]{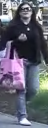}
\includegraphics[width=6mm]{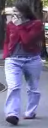}
\includegraphics[width=6mm]{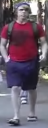}
\includegraphics[width=6mm]{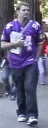}
\includegraphics[width=6mm]{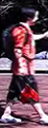}
\hspace{10mm}
\includegraphics[width=6mm]{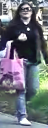}
\includegraphics[width=6mm]{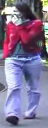}
\includegraphics[width=6mm]{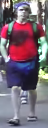}
\includegraphics[width=6mm]{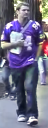}
\includegraphics[width=6mm]{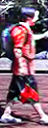}\\
\vspace{1mm}
\includegraphics[width=6mm]{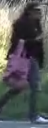}
\includegraphics[width=6mm]{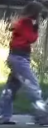}
\includegraphics[width=6mm]{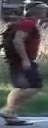}
\includegraphics[width=6mm]{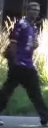}
\includegraphics[width=6mm]{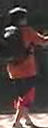}
\hspace{10mm}
\includegraphics[width=6mm]{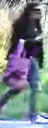}
\includegraphics[width=6mm]{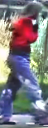}
\includegraphics[width=6mm]{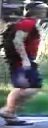}
\includegraphics[width=6mm]{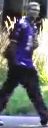}
\includegraphics[width=6mm]{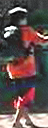}\\
(a) \hspace{35mm} (b)

\caption{(a) Example pairs of images from the VIPeR database \cite{gray2007evaluating}. (b) Processed images in (a) by Retinex. Images in the same column represent the same person.}
\label{fig:viper}
\end{figure}

Color is an important feature for describing person images. However, the illumination conditions across cameras can be very different, and the camera settings might also be different from camera to camera. Therefore, the perceived colors of the same person may vary largely from different camera views. For example, Fig. \ref{fig:viper} (a) shows some sample images from the VIPeR database \cite{gray2007evaluating}. It can be seen that images of the same person across the two camera views have a large difference in illumination and color appearance.

In this paper, we propose to apply the Retinex algorithm \cite{land1971lightness,jobson1997properties,jobson1997multiscale} to preprocess person images. Retinex considers human lightness and color perception. It aims at producing a color image that is consistent to human observation of the scene. The restored image usually contains vivid color information, especially enhanced details in shadowed regions.

We implement the multiscale Retinex algorithm according to \cite{jobson1997multiscale}, which combines the
small-scale Retinex for dynamic range compression and the large-scale Retinex for tonal rendition simultaneously. As a result, the algorithm handles both the color constancy and dynamic range compression automatically, achieving a good approximation to human visual perception. Specifically, we use two scales of center/surround Retinex, with $\sigma=5$ and $\sigma=20$. Besides, we automatically compute the gain/offset parameters so that the resulting intensities linearly stretches in [0,255].

Fig. \ref{fig:viper} (b) shows some examples of the processed images by our implementation of Retinex. Comparing to Fig. \ref{fig:viper} (a), it can be observed that the Retinex images of the same person across cameras have a better consistency in lighting and color. This makes person re-identification easier than using the original images. With the Retinex images, we apply the HSV color histogram to extract color features.
%
%\begin{figure}
%\centering
%\includegraphics[width=8mm]{retinex/106_0}
%\includegraphics[width=8mm]{retinex/110_0}
%\includegraphics[width=8mm]{retinex/118_0}
%\includegraphics[width=8mm]{retinex/120_0}
%\includegraphics[width=8mm]{retinex/353_90}
%\includegraphics[width=8mm]{retinex/692_0}\\
%\vspace{1mm}
%\includegraphics[width=8mm]{retinex/106_90}
%\includegraphics[width=8mm]{retinex/110_90}
%\includegraphics[width=8mm]{retinex/118_90}
%\includegraphics[width=8mm]{retinex/120_90}
%\includegraphics[width=8mm]{retinex/353_135}
%\includegraphics[width=8mm]{retinex/692_180}
%\caption{Processed images in Fig. \ref{fig:viper} by Retinex. Images in the same column represent the same person.}
%\label{fig:viper-retinex}
%\end{figure}
%

In addition to color description, we also apply the Scale Invariant Local Ternary Pattern (SILTP) \cite{Liao-CVPR-10} descriptor for illumination invariant texture description. SILTP is an improved operator over the well-known Local Binary Pattern (LBP) \cite{Ojala-PR-96}. In fact, LBP has a nice invariant property under monotonic gray-scale transforms, but it is not robust to image noises. SILTP improves LBP by introducing a scale invariant local comparison tolerance, achieving invariance to intensity scale changes and robustness to image noises.

\subsection{Dealing with Viewpoint Changes}
Pedestrians under different cameras usually appear in different viewpoint. For example, a person with frontal view in a camera may appear in back view under another camera. Therefore, matching persons in different viewpoints is also difficult. To address this, \cite{prosser2010person,zheng2011person} proposed to equally divide a person image into six horizontal stripes, and a single histogram is computed in each stripe. This feature has made a success in viewpoint invariant person representation \cite{prosser2010person,zheng2011person,Liu-ECCVW-2012}. However, it may also lose spatial details within a stripe, thus affecting its discriminative power.

\begin{figure}
\centering
\includegraphics[width=55mm]{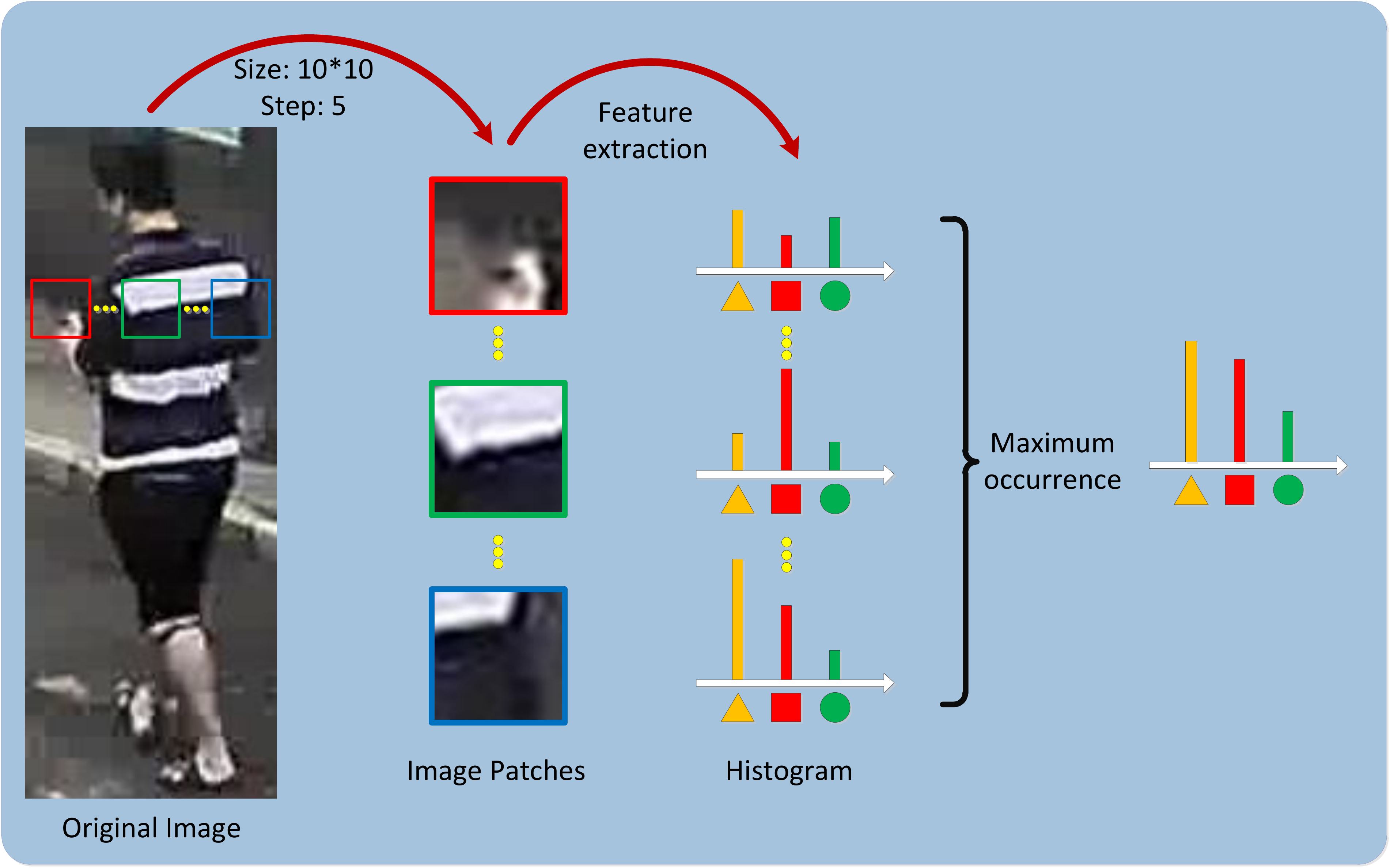}
\caption{Illustration of the LOMO feature extraction method.}
\label{fig:feature}
\end{figure}

We propose to use sliding windows to describe local details of a person image. Specifically, we use a subwindow size of $10\times10$, with an overlapping step of $5$ pixels to locate local patches in $128\times48$ images. Within each subwindow, we extract two scales of SILTP histograms (SILTP$_{4,3}^{0.3}$ and SILTP$_{4,5}^{0.3}$), and an $8\times8\times8$-bin joint HSV histogram. Each histogram bin represents the occurrence probability of one pattern in a subwindow. To address viewpoint changes, we check all subwindows at the same horizontal location, and maximize the local occurrence of each pattern (i.e. the same histogram bin) among these subwindows. The resulting histogram achieves some invariance to viewpoint changes, and at the same time captures local region characteristics of a person. Fig. \ref{fig:feature} shows the procedure of the proposed LOMO feature extraction.

To further consider the multi-scale information, we build a three-scale pyramid representation, which downsamples the original $128\times48$ image by two $2\times2$ local average pooling operations, and repeats the above feature extraction procedure. By concatenating all the computed local maximal occurrences, our final descriptor has $(8*8*8$ color bins + $3^4*2$ SILTP bins $)*(24 + 11 + 5$ horizontal groups $)=26,960$ dimensions. Finally, we apply a log transform to suppress large bin values, and normalize both HSV and SILTP features to unit length. Since we only use simple HSV and SILTP features, the proposed feature extraction method is efficient to compute (see Section \ref{subsubsec:time}).

\section{Cross-view Quadratic Discriminant Analysis}\label{sec:XQDA}
\subsection{Bayesian Face and KISSME Revisit}
Consider a sample difference $\Delta=\mathbf{x}_i - \mathbf{x}_j$. $\Delta$ is called the intrapersonal difference if $y_i=y_j$, while it is called the extrapersonal difference if $y_i\ne y_j$ \cite{Moghaddam-PR-2000}. Accordingly, two classes of variations can be defined: the intrapersonal variations $\Omega_I$ and the extrapersonal variations $\Omega_E$. Therefore, in this way the multi-class classification problem can be solved by distinguishing the above two classes. Moghaddam et al. \cite{Moghaddam-PR-2000} proposed to model each of the two classes with a multivariate Gaussian distribution. This corresponds to a QDA model with the defined $\Omega_I$ and $\Omega_E$ as two classes. Furthermore, it was noticed in \cite{Moghaddam-PR-2000} that both $\Omega_I$ and $\Omega_E$ have zero mean. The resulting algorithm is called Bayesian face applied to face recognition. Interestingly, in \cite{kostinger2012large}, K{\"o}stinger et al. also derived a similar approach called KISSME via the log likelihood ratio test of the two Gaussian distributions, and applied it to person re-identification.

Formally, the Bayesian face and the KISSME algorithms are formulated as follows. Under the zero-mean Gaussian distribution, the likelihoods of observing $\Delta$ in $\Omega_I$ and $\Omega_E$ are defined as
\begin{equation}
P(\Delta|\Omega_I)=\frac{1}{(2\pi)^{d/2} |\Sigma_I|^{1/2}} e^{-\frac{1}{2} \Delta^T \Sigma_I^{-1} \Delta},
\end{equation}
\begin{equation}
P(\Delta|\Omega_E)=\frac{1}{(2\pi)^{d/2} |\Sigma_E|^{1/2}} e^{-\frac{1}{2} \Delta^T \Sigma_E^{-1} \Delta},
\end{equation}
where $\Sigma_I$ and $\Sigma_E$ are the covariance matrices of $\Omega_I$ and $\Omega_E$, respectively, and $n_I$ and $n_E$ denotes the number of samples in the two classes. %According to the Bayesian rule, the posterior probability of assigning an observation $\Delta$ to $\Omega_I$ and $\Omega_E$ are $P(\Omega_I | \Delta) = \frac{P(\Delta|\Omega_I) P(\Omega_I)}{P(\Delta)}$ and $P(\Omega_E | \Delta) = \frac{P(\Delta|\Omega_E) P(\Omega_E)}{P(\Delta)}$, respectively. Therefore, the decision function is obtained by comparing $P(\Omega_E | \Delta)$ and $P(\Omega_I | \Delta)$. By doing a log-likelihood ratio test, the decision function can be simplified as
By applying the Bayesian rule and the log-likelihood ratio test, the decision function can be simplified as
\begin{equation}
f(\Delta) = \Delta^T (\Sigma_I^{-1} - \Sigma_E^{-1}) \Delta,
\end{equation}
%
%By applying the Bayesian rule and the log-likelihood ratio test, the distance function can be simplified as
and so the derived distance function between $\mathbf{x}_i$ and $\mathbf{x}_j$ is
\begin{equation}\label{equ:distance}
d(\mathbf{x}_i, \mathbf{x}_j) = (\mathbf{x}_i - \mathbf{x}_j)^T (\Sigma_I^{-1} - \Sigma_E^{-1}) (\mathbf{x}_i - \mathbf{x}_j).
\end{equation}
Therefore, learning the distance function corresponds to estimating the covariant matrices $\Sigma_I$ and $\Sigma_E$.

\subsection{XQDA}
Usually, the original feature dimensions $d$ is large, and a low dimensional space $\mathds{R}^r$ ($r<d$) is preferred for classification. \cite{Moghaddam-PR-2000} suggested to decompose $\Sigma_I$ and $\Sigma_E$ separately to reduce the dimensions. In \cite{kostinger2012large}, PCA was applied, then $\Sigma_I$ and $\Sigma_E$ were estimated in the PCA subspace. However, both methods are not optimal because the dimension reduction does not consider the distance metric learning.

In this paper, we extend the Bayesian face and KISSME approaches to cross-view metric learning, where we consider to learn a subspace $W=(\mathbf{w}_1, \mathbf{w}_2, \ldots, \mathbf{w}_r) \in \mathds{R}^{d\times r}$ with cross-view data, and at the same time learn a distance function in the $r$ dimensional subspace for the cross-view similarity measure. Suppose we have a cross-view training set $\{\mathbf{X}, \mathbf{Z}\}$ of $c$ classes, where $\mathbf{X} = (\mathbf{x}_1, \mathbf{x}_2, \ldots, \mathbf{x}_n) \in \mathds{R}^{d\times n}$ contains $n$ samples in a $d$-dimensional space from one view, $\mathbf{Z} = (\mathbf{z}_1, \mathbf{z}_2, \ldots, \mathbf{z}_m) \in \mathds{R}^{d\times m}$ contains $m$ samples in the same $d$-dimensional space but from the other view. The cross-view matching problem arises from many applications, like heterogeneous face recognition \cite{Liao-AMFG-2009} and viewpoint invariant person re-identification \cite{Gray-ECCV-08}.
Note that $\mathbf{Z}$ is the same with $\mathbf{X}$ in the single-view matching scenario.

Considering a subspace $W$, the distance function Eq. (\ref{equ:distance}) in the $r$ dimensional subspace is computed as
\begin{equation}\label{equ:distance-W}
d_W(\mathbf{x}, \mathbf{z}) = (\mathbf{x} - \mathbf{z})^T W (\Sigma_I'^{-1} - \Sigma_E'^{-1}) W^T (\mathbf{x} - \mathbf{z}),
\end{equation}
where $\Sigma_I'=W^T \Sigma_I W$ and $\Sigma_E'= W^T \Sigma_E W$.
\begin{figure}
\centering
\includegraphics[width=30mm]{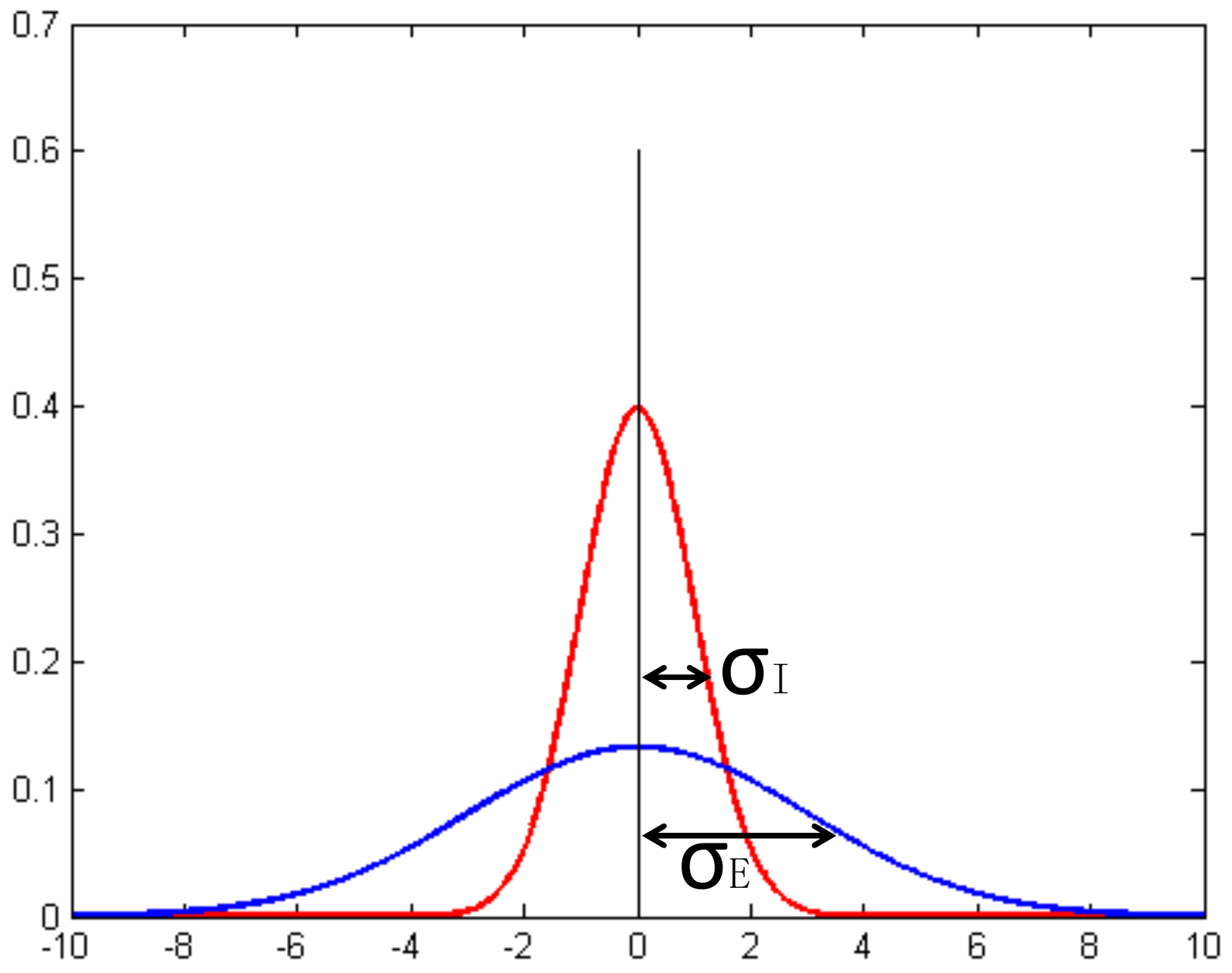}
\caption{Distributions of $\Omega_I$ and $\Omega_E$ in one projected dimension.}
\label{fig:gauss}
\end{figure}
Therefore, we needs to learn a kernel matrix $M(W) = W (\Sigma_I'^{-1} - \Sigma_E'^{-1}) W^T$. However, directly optimizing $d_W$ is difficult because $W$ is contained in two inverse matrices.

Recall that $\Omega_I$ and $\Omega_E$ have zero mean, then given a basis $\mathbf{w}$, the projected samples of the two classes will still center at zero, but may have different variances, as shown in Fig. \ref{fig:gauss}. In this case, the traditional Fisher criterion used to derive LDA is no longer suitable because the two classes have the same mean. However, the variances $\sigma_I$ and $\sigma_E$ can still be used to distinguish the two classes. Therefore, we can optimize the projection direction $\mathbf{w}$ such that $\sigma_E(\mathbf{w}) / \sigma_I(\mathbf{w})$ is maximized. Notice that $\sigma_I(\mathbf{w}) = \mathbf{w}^T \Sigma_I \mathbf{w}$ and $\sigma_E(\mathbf{w}) = \mathbf{w}^T \Sigma_E \mathbf{w}$, therefore, the objective $\sigma_E(\mathbf{w}) / \sigma_I(\mathbf{w})$ corresponds to the Generalized Rayleigh Quotient
\begin{equation}
J(\mathbf{w}) = \frac{\mathbf{w}^T \Sigma_E \mathbf{w}}{\mathbf{w}^T \Sigma_I \mathbf{w}}.
\end{equation}
The maximization of $J(\mathbf{w})$ is equivalent to
\begin{equation}
\max_\mathbf{w} \mathbf{w}^T \Sigma_E \mathbf{w}, ~s.t.~\mathbf{w}^T \Sigma_I \mathbf{w} = 1,
\end{equation}
which can be solved by the generalized eigenvalue decomposition problem as similar in LDA. That is, the largest eigenvalue of $\Sigma_I^{-1} \Sigma_E$ is the maximum value of $J(\mathbf{w})$, and the corresponding eigenvector $\mathbf{w}_1$ is the solution. Furthermore, the solution orthogonal to $\mathbf{w}_1$ and corresponding to the second largest value of $J(\mathbf{w})$ is the eigenvector of the second largest eigenvalue of $\Sigma_I^{-1} \Sigma_E$, and so on. Therefore, with $W=(\mathbf{w}_1, \mathbf{w}_2, \ldots, \mathbf{w}_r)$ we learn a discriminant subspace, as well as a distance function in the learned subspace, as defined in Eq. (\ref{equ:distance-W}). We call the derived algorithm Cross-view Quadratic Discriminant Analysis (XQDA) to reflect its connection to QDA and the output of a cross-view metric.

\subsection{Practical Computation}
The computation of the two covariance matrices $\Sigma_I$ and $\Sigma_E$ require $O(N k d^2)$ and $O(n m d^2)$ multiplication operations, respectively, where $N=max(m,n)$, and $k$ represents the average number of images in each class. To reduce the computation, we show that

\begin{equation}
n_I \Sigma_I = \widetilde{\mathbf{X}} \widetilde{\mathbf{X}}^T + \widetilde{\mathbf{Z}} \widetilde{\mathbf{Z}}^T
-\mathbf{S} \mathbf{R}^T -\mathbf{R} \mathbf{S}^T,
\end{equation}
where $\widetilde{\mathbf{X}} = (\sqrt{m_1} \mathbf{x}_1, \sqrt{m_1} \mathbf{x}_2, \ldots, \sqrt{m_1} \mathbf{x}_{n_1}, \ldots, \sqrt{m_c} \mathbf{x}_n)$, $\widetilde{\mathbf{Z}} = (\sqrt{n_1} \mathbf{z}_1, \sqrt{n_1} \mathbf{z}_2, \ldots, \sqrt{n_1} \mathbf{z}_{m_1}, \ldots, \sqrt{n_c} \mathbf{z}_m)$, $\mathbf{S} = (\sum_{y_i=1} \mathbf{x}_i, \sum_{y_i=2} \mathbf{x}_i, \ldots, \sum_{y_i=c} \mathbf{x}_i)$, $\mathbf{R} = (\sum_{l_j=1} \mathbf{z}_j, \sum_{l_j=2} \mathbf{z}_j, \ldots, \sum_{l_j=c} \mathbf{z}_j)$, $y_i$ and $l_j$ are class labels, $n_k$ is the number of samples in class $k$ of $\mathbf{X}$, and $m_k$ is the number of samples in class $k$ of $\mathbf{Z}$. Besides,
\begin{equation}
n_E \Sigma_E = m \mathbf{X} \mathbf{X}^T + n \mathbf{Z} \mathbf{Z}^T - \mathbf{s} \mathbf{r}^T - \mathbf{r} \mathbf{s}^T - n_I \Sigma_I,
\end{equation}
where $\mathbf{s} = \sum_{i=1}^n \mathbf{x}_i$ and $\mathbf{r} = \sum_{j=1}^m \mathbf{z}_j$. The above simplification shows that the computations of $\Sigma_I$ and $\Sigma_E$ are both reduced to $O(N d^2)$. It can be observed that, $\Sigma_I$ and $\Sigma_E$ can be computed directly from sample mean and covariance of each class and all classes, so there is no need to actually compute the $mn$ pairs of sample differences required in many other metric learning algorithms.

Another practical issue is that, $\Sigma_I$ may be singular, resulting that $\Sigma_I^{-1}$ cannot be computed. Therefore, it is useful to add a small regularizer to the diagonal elements of $\Sigma_I$, as usually done in similar problems like LDA. This will make the estimation of $\Sigma_I$ more smooth and robust. Empirically we find that, when all samples are normalized to unit length, a value of $0.001$ as a regularizer can be commonly applied to improve the result.

Finally, there is a remaining issue of selecting the dimensionality of the derived XQDA subspace. In real applications, there should be a consideration to have a low dimensional subspace to ensure the processing speed. Beyond this consideration, we find that having the selected eigenvalues of $\Sigma_I^{-1} \Sigma_E$ larger than 1 is a useful signature to determine the dimensions. This is because the eigenvalue of $\Sigma_I^{-1} \Sigma_E$ corresponds to $\sigma_E / \sigma_I$ in Fig. \ref{fig:gauss}, and $\sigma_E < \sigma_I$ may not provide useful discriminant information.

\section{Experiments}\label{sec:exp}
%The proposed algorithm was evaluated on four challenging person re-identification databases, VIPeR \cite{gray2007evaluating}, QMUL GRID \cite{loy2009multi}, CUHK Campus \cite{Li-ACCV-2012}, and CUHK03 \cite{Li-CVPR-2014-DeepReID}. Two available features provided by \cite{kostinger2012large} and \cite{loy2009multi} for person re-identification were evaluated and compared to the proposed feature. Several state-of-the-art metric learning algorithms with the same feature representation were compared, and the state-of-the-art published results on the four datasets were also compared. We detail the experimental description below.

\subsection{Experiments on VIPeR}
VIPeR \cite{gray2007evaluating} is a challenging person re-identification database that has been widely used for benchmark evaluation. It contains 632 pairs of person images, captured by a pair of cameras in an outdoor environment. Images in VIPeR contains large variations in background, illumination, and viewpoint. Fig. \ref{fig:viper} (a) shows some example pairs of images from the VIPeR database. All images are scaled to $128\times48$ pixels. The widely adopted experimental protocol on this database is to randomly divide the 632 pairs of images into half for training and the other half for testing, and repeat the procedure 10 times to get an average performance. We followed this procedure in our experiments.

\subsubsection{Comparison of Metric Learning Algorithms}
\begin{figure}
\centering
\includegraphics[width=40mm]{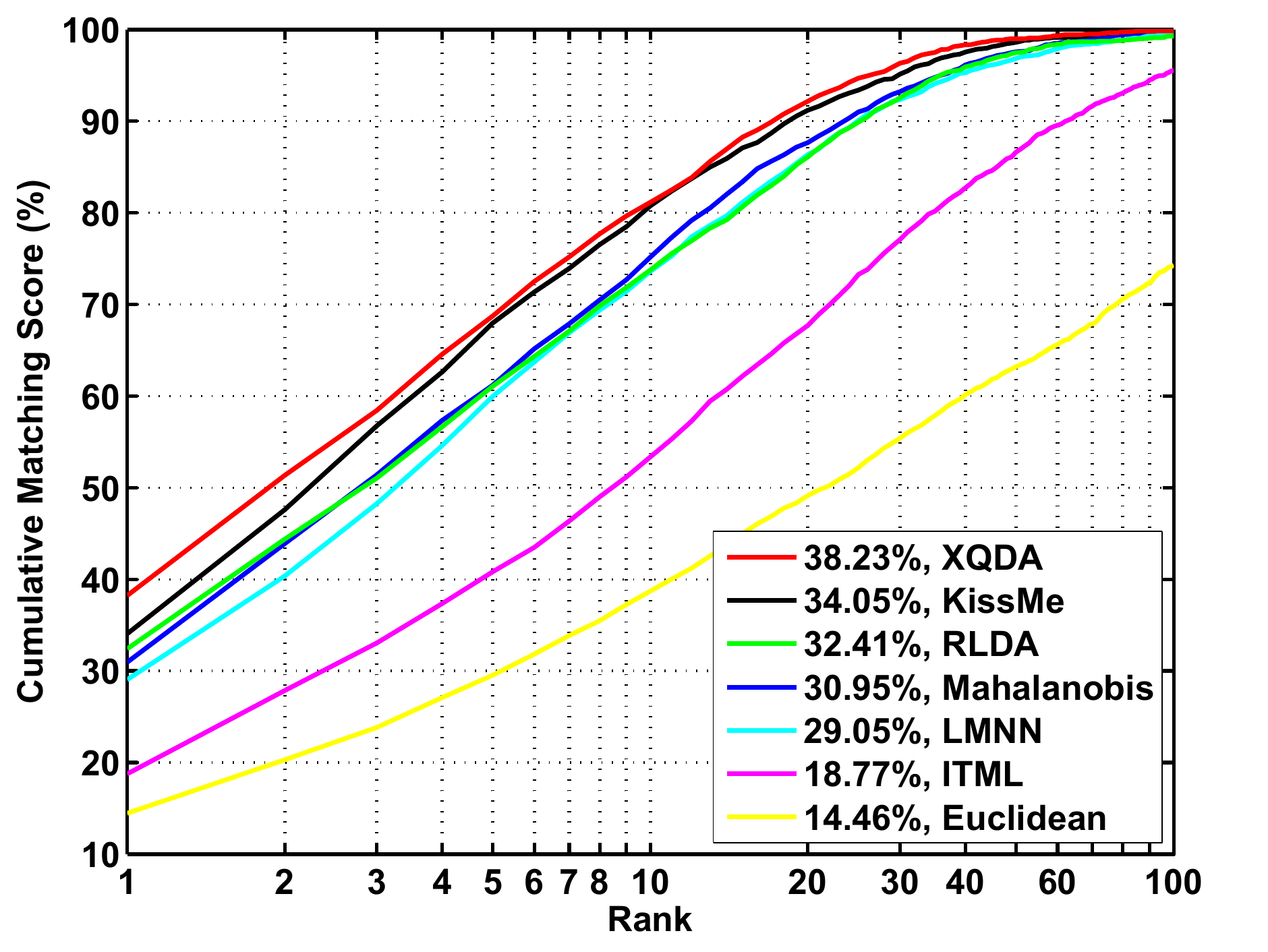}
\includegraphics[width=40mm]{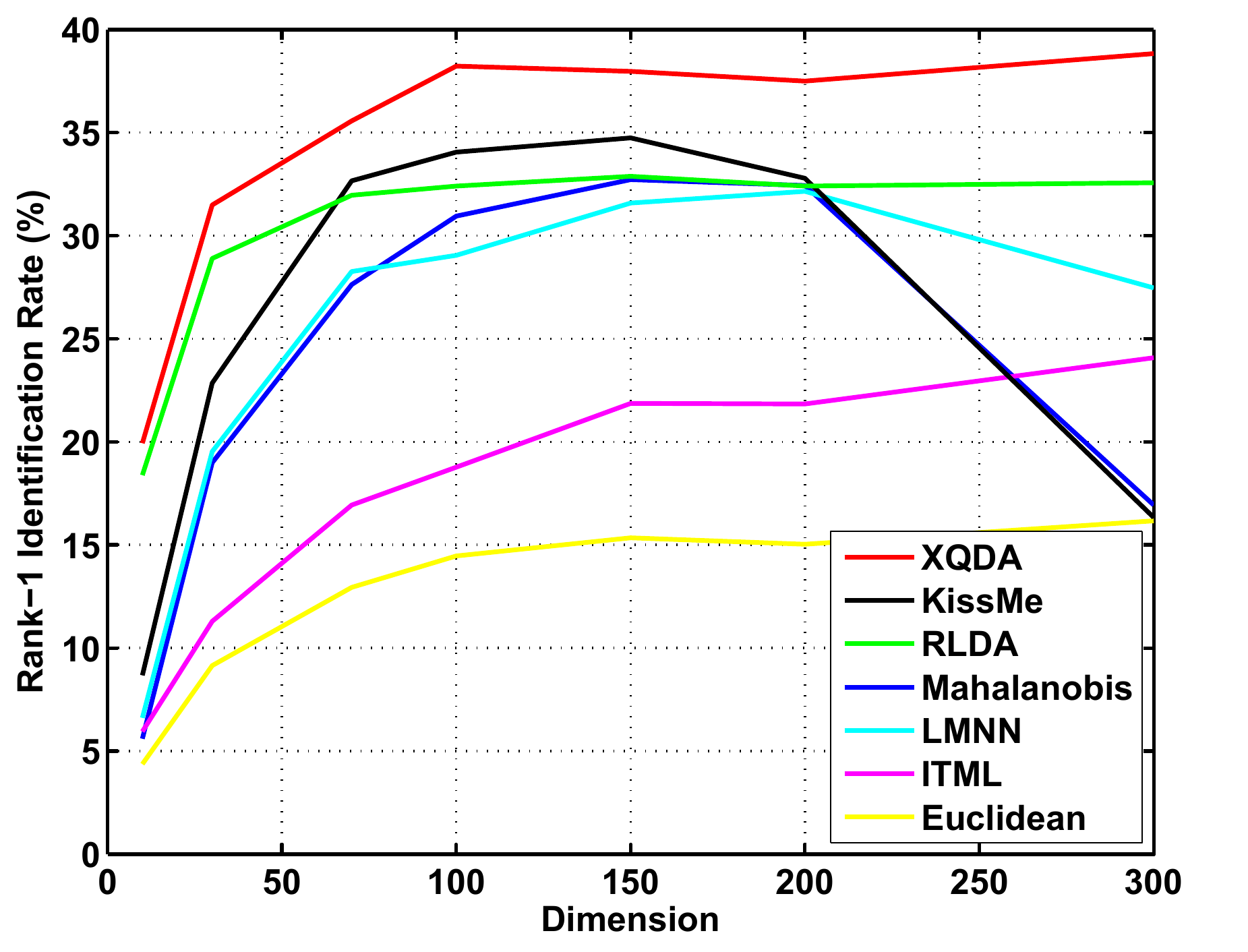}
(a) \hspace{35mm} (b)
\caption{Comparison of metric learning algorithms with the same LOMO feature on the VIPeR database \cite{gray2007evaluating} (P=316). (a) CMC curves with feature reduced to 100 dimensions. (b) Rank-1 identification rates with varying subspace dimensions.}
\label{fig:viper-lomo}
\end{figure}

We evaluated the proposed XQDA algorithm and several metric learning algorithms, including Euclidean distance, Mahalanobis distance trained with genuine pairs\cite{kostinger2012large}, LMNN v2.5\cite{weinberger2006distance}, ITML \cite{davis2007information}, KISSME \cite{kostinger2012large}, and RLDA \cite{Ye-CIKM-06}, with the same LOMO feature. %A feature set extracted from the VIPeR database is also provided by \cite{kostinger2012large}, which is a mixture of HSV, Lab, and LBP features. We first used this public feature set to evaluate the proposed XQDA algorithm, with a comparison to the above metric learning algorithms. For the compared algorithms, according to \cite{kostinger2012large}, PCA was first applied to reduce the feature dimensionality to 34.
%We first used the LOMO feature to evaluate the proposed XQDA algorithm, with a comparison to the above metric learning algorithms.
For the compared algorithms, PCA was first applied to reduce the feature dimensionality to 100.The proposed XQDA algorithm and RLDA also learned a 100-dimensional subspace. The resulting Cumulative Matching Characteristic (CMC) curves are shown in Fig. \ref{fig:viper-lomo} (a). It can be seen that the proposed method is better than the compared metric learning algorithms. This indicates that XQDA successfully learns a discriminant subspace as well as an effective metric. Besides, we also investigate how the performance varies with different subspace dimensions, as shown in Fig. \ref{fig:viper-lomo} (b). It can be observed that XQDA consistently performs the best with all dimensions.

\subsubsection{Comparison of Features}
\begin{figure}
\centering
\includegraphics[width=40mm]{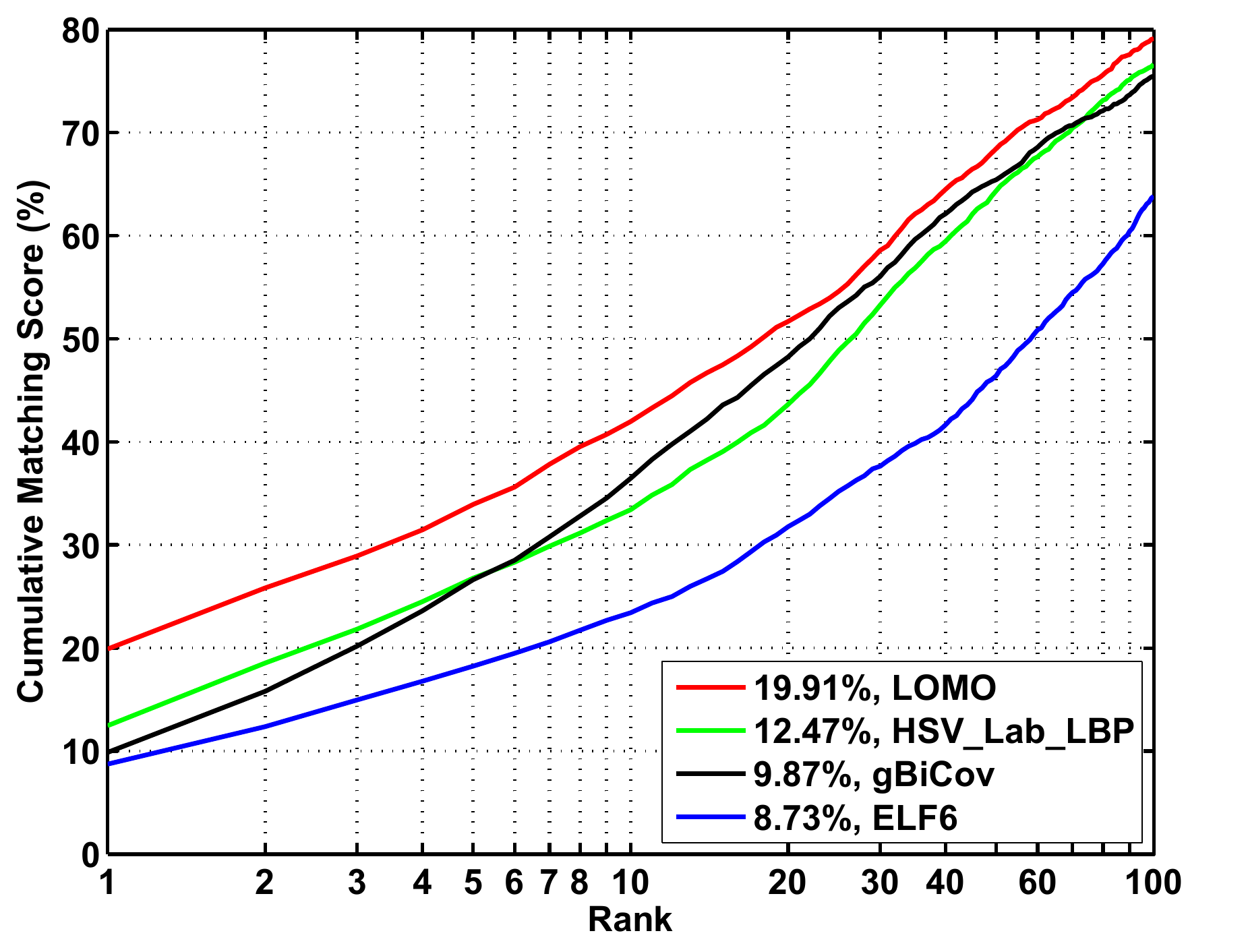}
\includegraphics[width=40mm]{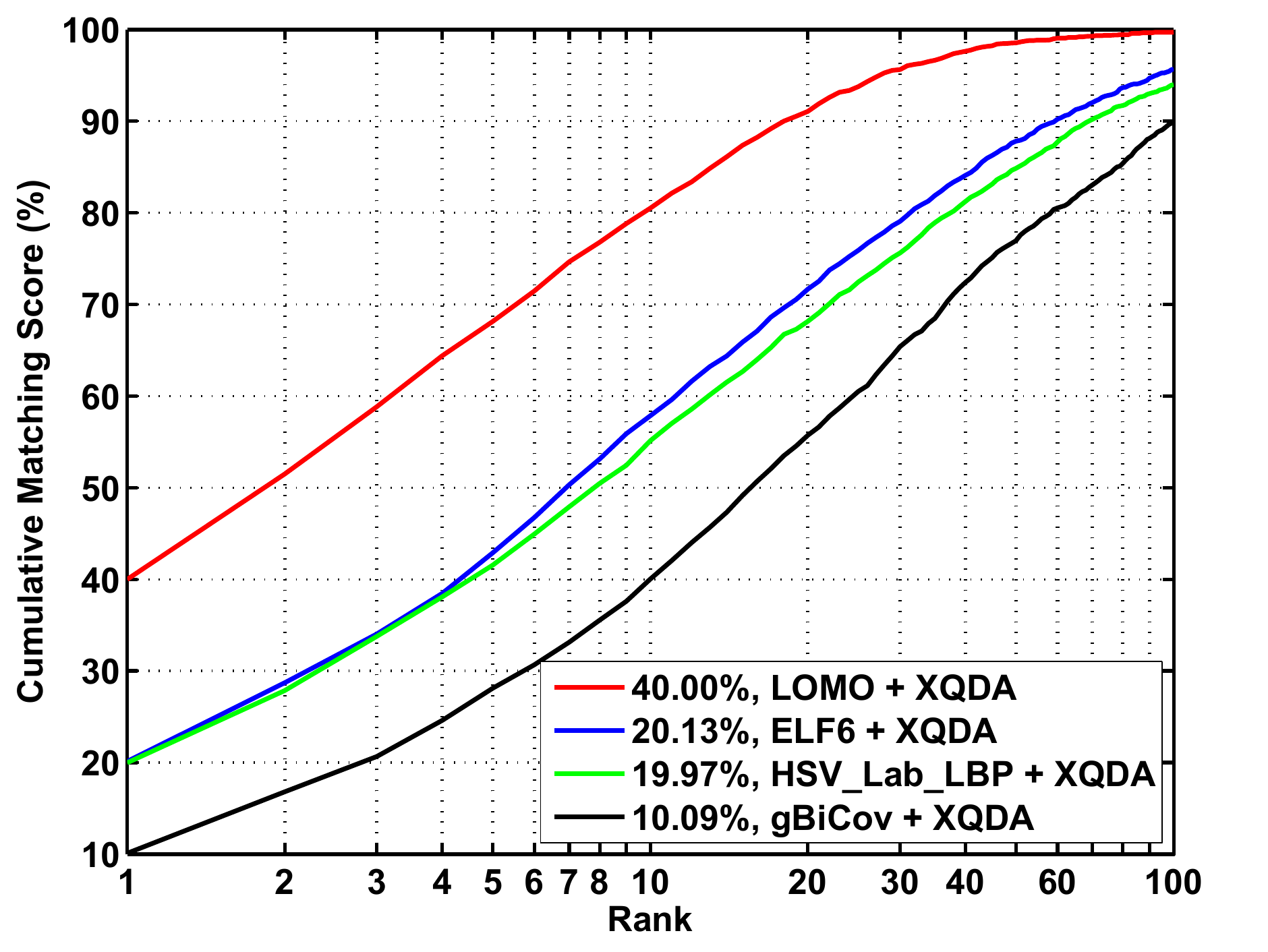}
(a) Cosine \hspace{25mm} (b) XQDA
\caption{CMC curves and rank-1 identification rates on the VIPeR database \cite{gray2007evaluating} (P=316), by comparing the proposed LOMO feature to three available features, ELF6 \cite{Zheng-PAMI-2013}, HSV+Lab+LBP \cite{kostinger2012large}, and gBiCov \cite{Ma-IVC-2014-kBiCov}.}
\label{fig:viper-feature}
\end{figure}

Next, we compared the proposed LOMO feature with other three available person re-identification features. The first feature is called Ensemble of Local Features (ELF), proposed in \cite{Gray-ECCV-08}, and later modified by \cite{prosser2010person,zheng2011person}. We used the implementation in \cite{Zheng-PAMI-2013}, denoted by ELF6, which is computed from histograms in six equally divided horizontal stripes. Eight color channels (RGB, HSV, and YCbCr) and 21 texture filters (8 Gabor filters and 13 Schmid filters) are used for the histogram representation. %, resulting in a 2784-dimensional feature vector for each image.
%This feature is considered to be effective for person re-identification and has demonstrated good performance in several papers \cite{prosser2010person,zheng2011person,Liu-ECCVW-2012,loy2013person,Zheng-PAMI-2013,Ma-TIP-2014-MtMCML}.
The other feature is proposed in \cite{kostinger2012large}, which applied the HSV, and Lab color feature, as well as a texture feature extracted by LBP. The third feature called gBiCov\footnote{We used the author's implementation (available in \url{http://vipl.ict.ac.cn/members/bpma}) and the default parameters, which may not reflect the best status.} \cite{Ma-IVC-2014-kBiCov} is a combination of Biologically
Inspired Features (BIF) and Covariance descriptors. We applied both the direct Cosine similarity measure and the XQDA algorithm to compare the four different kinds of features, resulting in the CMC curves shown in Fig. \ref{fig:viper-feature}. For consistency, in the following experiments we determined the subspace dimensions of XQDA automatically by accepting all eigenvalues of $\Sigma_I^{-1} \Sigma_E$ that are larger than 1, as discussed earlier. From Fig. \ref{fig:viper-feature} (a) it can be seen that the raw LOMO feature outperforms the other existing features. What's more, Fig. \ref{fig:viper-feature} (b) shows that the performance improvement is more significant with the help of XQDA. Since these kinds of features are similar in fusing color and texture information, the improvement made by the proposed LOMO feature is mainly due to the specific consideration of handling illumination and viewpoint changes.
%
%\begin{figure}
%\centering
%\includegraphics[width=45mm]{viper/viper_xqda}
%\caption{CMC curves and rank-1 identification rates on the VIPeR database \cite{gray2007evaluating} (P=316), by comparing the proposed LOMO feature to two available features, ELF6 \cite{Zheng-PAMI-2013} and HSV+Lab+LBP \cite{kostinger2012large}.}
%\label{fig:viper-xqda}
%\end{figure}
%
%

\subsubsection{Comparison to the State of the Art}
\begin{figure}
\centering
\includegraphics[width=55mm]{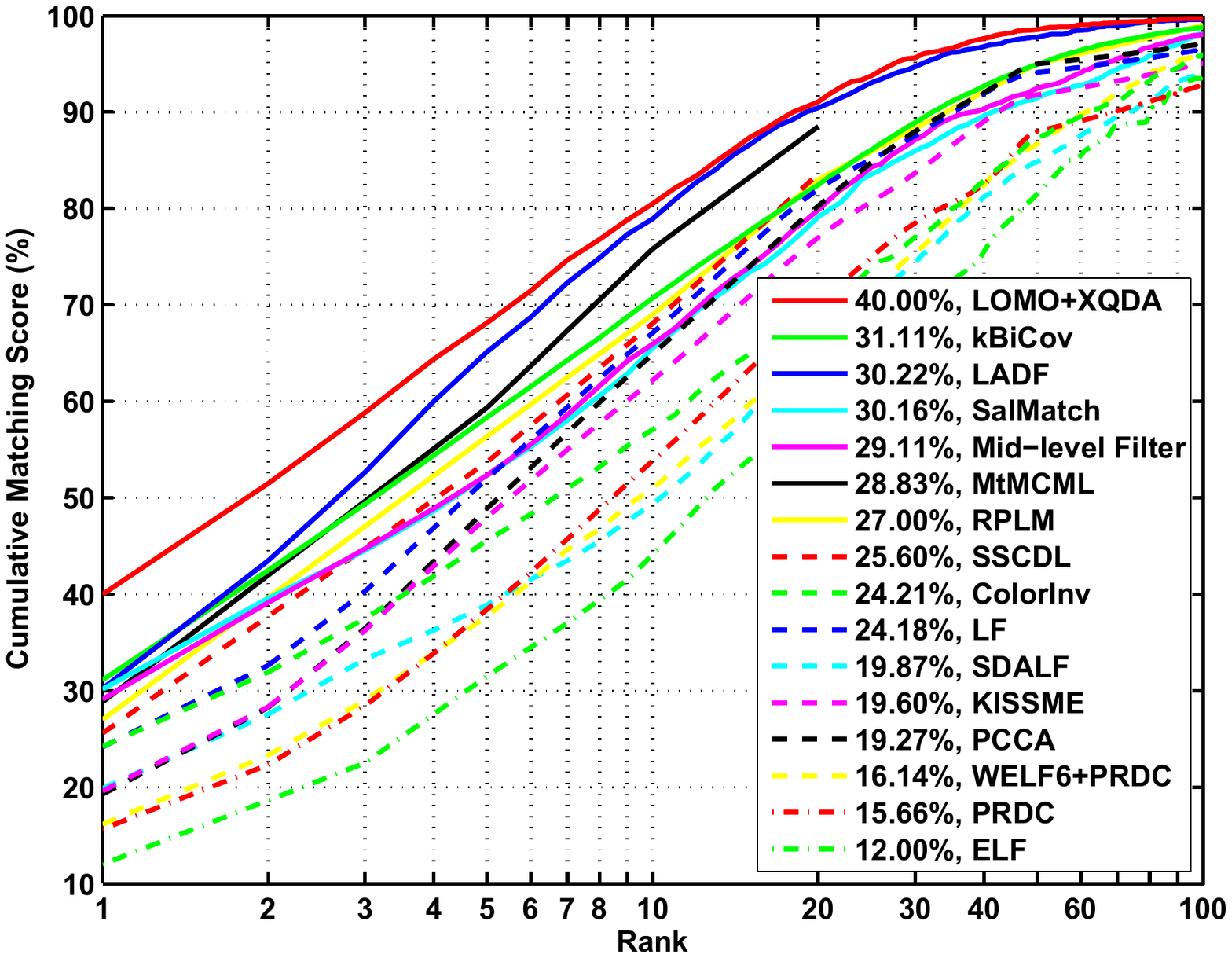}
\caption{CMC curves and rank-1 identification rates on the VIPeR database \cite{gray2007evaluating} (P=316) by comparing the proposed LOMO+XQDA method to other state of the art algorithms.}
\label{fig:viper-cmc}
\end{figure}

Finally, we compare the performance of the proposed approach to the state-of-the-art results reported on the VIPeR database, which are summarized in Fig. \ref{fig:viper-cmc} and Table \ref{tab:viper-best}. Four methods, the SCNCD \cite{Yang-ECCV-2014}, kBiCov \cite{Ma-IVC-2014-kBiCov}, LADF \cite{li2013learning}, and SalMatch \cite{zhao2013person} report the best performances on the VIPeR dataset to date, which exceed 30\% at rank 1. From Table \ref{tab:viper-best} it can be observed that the proposed algorithm achieves the new state of the art, 40\% at rank 1, outperforming the second best one SCNCD by 2.2\%.
\begin{table}
    \centering
    \caption{Comparison of state-of-the-art results reported with the VIPeR database (P=316). The cumulative matching scores (\%) at rank 1, 10, and 20 are listed.}\label{tab:viper-best}
    \begin{tabular}{|@{}c@{} | @{ }c@{} | @{ }c@{} | @{ }c@{} | @{}c@{}|}
        \hline
         \textbf{Method} & \textbf{rank=1} & \textbf{rank=10} & \textbf{rank=20} & \textbf{Reference}\\
        \hline\hline
        LOMO+XQDA & \textbf{40.00} & 80.51 & \textbf{91.08} & Proposed\\
        SCNCD & 37.80 & \textbf{81.20} & 90.40 & 2014 ECCV \cite{Yang-ECCV-2014}\\
        kBiCov & 31.11 & 70.71 & 82.45 & 2014 IVC \cite{Ma-IVC-2014-kBiCov}\\
        LADF & 30.22 & 78.92 & 90.44 & 2013 CVPR \cite{li2013learning}\\
        SalMatch & 30.16 & 65.54 & 79.15 & 2013 ICCV \cite{zhao2013person}\\
        Mid-level Filter$^*$ & 29.11 & 65.95 & 79.87 & 2014 CVPR \cite{Zhao-CVPR-2014-MidLF}\\
        MtMCML & 28.83 & 75.82 & 88.51 & 2014 TIP \cite{Ma-TIP-2014-MtMCML}\\
        RPLM & 27.00 & 69.00 & 83.00 & 2012 ECCV \cite{Hirzer-ECCV-2012-RPLM}\\
        LDFV & 26.53 & 70.88 & 84.63 & 2012 ECCVW \cite{ma2012local}\\
        SSCDL & 25.60 & 68.10 & 83.60 & 2014 CVPR \cite{Liu-CVPR-2014-SSCDL}\\
        ColorInv & 24.21 & 57.09 & 69.65 & 2013 TPAMI \cite{Kvia-PAMI-2013-ColorInv}\\
        LF & 24.18 & 67.12 & 82.00 & 2013 CVPR \cite{Pedagadi-CVPR-2013-LF}\\
        SDALF & 19.87 & 49.37 & 65.73 & 2013 CVIU \cite{Bazzani-CVIU-2013-SDALF}\\
        KISSME & 19.60 & 62.20 & 77.00 & 2012 CVPR \cite{kostinger2012large}\\
        PCCA & 19.27 & 64.91 & 80.28 & 2012 CVPR \cite{Mignon-CVPR-2012-PCCA}\\
        WELF6+PRDC & 16.14 & 50.98 & 65.95 & 2012 ECCVW \cite{Liu-ECCVW-2012}\\
        PRDC & 15.66 & 53.86 & 70.09 & 2013 TPAMI \cite{Zheng-PAMI-2013}\\
        ELF & 12.00 & 44.00 & 61.00 & 2008 ECCV \cite{Gray-ECCV-08}\\
        \hline
    \end{tabular}\\
    \vspace{2mm}
    \scriptsize{$^*$Note that \cite{Zhao-CVPR-2014-MidLF} reports a 43.39\% rank-1 accuracy by fusing their method with LADF \cite{li2013learning}. Fusing different methods generally improves the performance. In fact, we also tried to fuse our method with LADF, and got a 50.32\% rank-1 identification rate.}
\end{table}

\subsection{Experiments on QMUL GRID}
\begin{figure}
\centering
\begin{tabular}{c c c c c c}
\includegraphics[height=16mm]{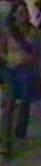}&
\includegraphics[height=16mm]{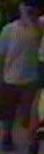}&
\includegraphics[height=16mm]{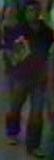}&
\includegraphics[height=16mm]{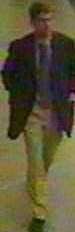}&
\includegraphics[height=16mm]{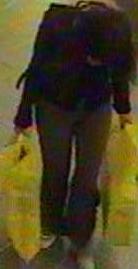}&
\includegraphics[height=16mm]{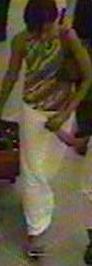}\\
%\vspace{1mm}
\includegraphics[height=16mm]{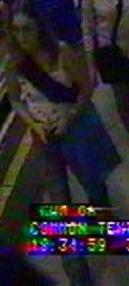}&
\includegraphics[height=16mm]{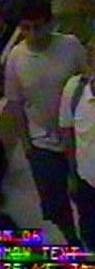}&
\includegraphics[height=16mm]{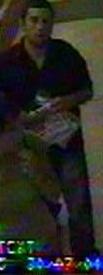}&
\includegraphics[height=16mm]{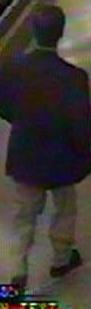}&
\includegraphics[height=16mm]{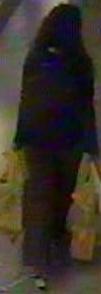}&
\includegraphics[height=16mm]{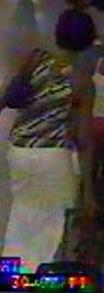}\\
\end{tabular}
\caption{Example pairs of images from the GRID database \cite{loy2009multi}. Images in the same column represent the same person.}
\label{fig:grid}
\end{figure}

The QMUL underGround Re-IDentification (GRID) dataset \cite{loy2009multi} is another challenging person re-identification test bed but have not been largely noticed. The GRID dataset was captured from 8 disjoint camera views in a underground station. It contains 250 pedestrian image pairs, with each pair contains two images of the same person from different camera views. Besides, there are 775 additional images that do not belong to the 250 persons which can be used to enlarge the gallery set. Sample images from GRID can be found in Fig. \ref{fig:grid}. It can be seen that these images have poor image quality and low resolutions, and contain large variations of illumination and viewpoint.

An experimental setting of 10 random trials is provided for the GRID dataset. For each trial, 125 image pairs are used for training, and the remaining 125 image pairs, as well as the 775 background images are used for test. The ELF6 feature set described in \cite{Liu-ECCVW-2012} is provided for developing machine learning algorithms.

We first applied the proposed method on the provided feature set of GRID. This leads to results of ``ELF6+XQDA'' listed in Table \ref{tab:grid-independent}. We compared available results from \cite{loy2013person} where the same feature set was used. Results shown in Table \ref{tab:grid-independent} indicates that the proposed joint dimension reduction and metric learning approach outperforms other distance learning algorithms such as RankSVM \cite{prosser2010person}, PRDC \cite{zheng2011person}, and MRank \cite{loy2013person}, except that the rank-1 accuracy of XQDA is slightly worse than MRank-RankSVM.
\begin{table}
\centering
\caption{Comparison of state-of-the-art results on the GRID database (P=900) without camera network information. Red and blue numbers are the best and second best results, respectively.}\label{tab:grid-independent}
\begin{tabular}{|@{}c@{}|@{}c@{}|@{}c@{}|@{}c@{}|}
  \hline	
  \textbf{Method} & \textbf{rank=1} & \textbf{rank=10} & \textbf{rank=20}\\
  \hline
ELF6 + L1-norm \cite{loy2013person}	        &4.40	&16.24	&24.80\\
ELF6 + RankSVM \cite{prosser2010person}	        &10.24 	&33.28  &43.68\\
ELF6 + PRDC	 \cite{Zheng-PAMI-2013}        &9.68 	&32.96 	&44.32\\
ELF6 + MRank-RankSVM \cite{loy2013person}	&\textcolor{blue}{12.24} 	&36.32  &46.56\\
ELF6 + MRank-PRDC \cite{loy2013person}	    &11.12  &35.76  &46.56 \\
ELF6 + XQDA &10.48    &\textcolor{blue}{38.64}   &\textcolor{red}{\textbf{52.56}}\\
LOMO + XQDA &\textcolor{red}{\textbf{16.56}}    &\textcolor{red}{\textbf{41.84}}   &\textcolor{blue}{52.40}\\
  \hline
\end{tabular}
\end{table}
\begin{table}
\centering
\caption{Comparison of state-of-the-art results on the GRID database (P=900) with camera network information. Red and blue numbers are the best and second best results, respectively.}\label{tab:grid-dependent}
\begin{tabular}{|@{}c@{}|@{}c@{}|@{}c@{}|@{}c@{}|}
  \hline	
  \textbf{Method} & \textbf{rank=1} & \textbf{rank=10} & \textbf{rank=20}\\
  \hline
ELF6 + MtMCML \cite{Ma-TIP-2014-MtMCML} &14.08  &\textcolor{blue}{45.84} 	&\textcolor{blue}{59.84}\\
ELF6 + XQDA &\textcolor{blue}{16.32}   &40.72    &51.76\\
LOMO + XQDA &\textcolor{red}{\textbf{18.96}}     &\textcolor{red}{\textbf{52.56}}     &\textcolor{red}{\textbf{62.24}}\\
  \hline
\end{tabular}
\end{table}

We also tried the proposed feature extraction method, and applied the same XQDA algorithm for metric learning. This corresponds to the results of the last row in Table \ref{tab:grid-independent}. The comparison shows that the new feature improves the performance at rank 1-10. Especially, a 4.32\% performance gain can be obtained for the rank-1 accuracy. This indicates that the new feature helps to reduce intra-class variations, so that the same person can be recognized at a higher rank.

Note that the above methods all trained a general model independent of camera views. A research in \cite{Ma-TIP-2014-MtMCML} show that the performance can be improved by utilizing the camera network information. Namely, their method MtMCML trained various metrics, each for a given camera view pair. We also followed this approach and trained several metrics depending on known camera pairs. Results listed in Table \ref{tab:grid-dependent} show that, while with the ELF6 feature the proposed method only improves the rank-1 accuracy over MtMCML, with the new LOMO feature the proposed method is clearly better than MtMCML. However, in practice we do not suggest this way of training because the camera views under evaluation are usually unseen, and it is not easy to label data for new camera views to retrain the algorithm.% In the following Subsection \ref{subsec:cuhk02} with the CUHK02 database, we will see that test on unseen camera views is more challenging than on learned camera views.

\subsection{Experiments on CUHK Campus}
\begin{figure}
\centering
\includegraphics[width=50mm]{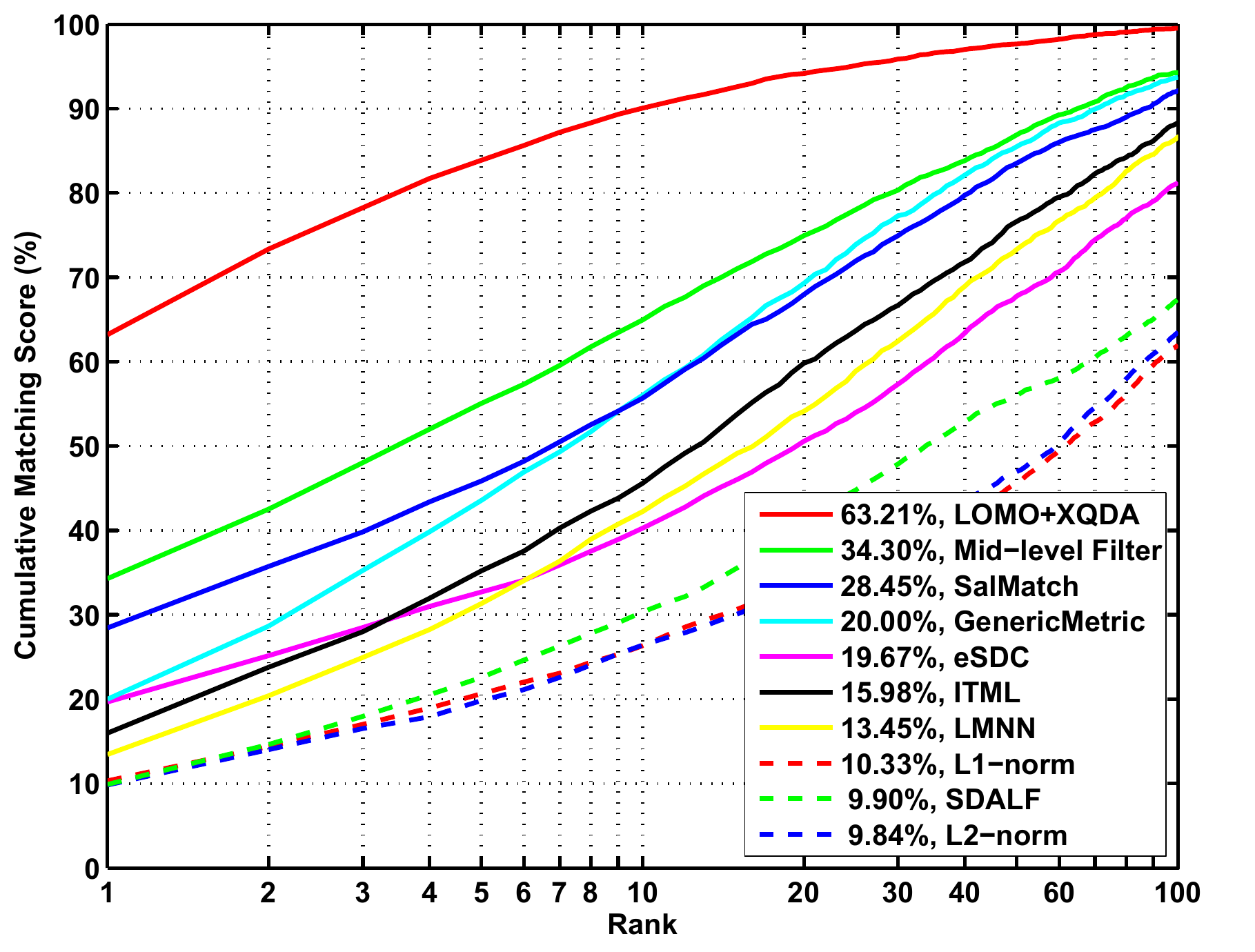}
\caption{Multi-shot CMC curves and rank-1 identification rates on the CUHK Campus database \cite{Li-ACCV-2012} (P=486, M=2). The compared results are from \cite{Zhao-CVPR-2014-MidLF}.}
\label{fig:cuhk01}
\end{figure}

The CUHK Campus dataset %\footnote{The CUHK01 dataset was previously called the CUHK Campus dataset. The CUHK01, CUHK02, and CUHK03 datasets can be downloaded from \url{http://www.ee.cuhk.edu.hk/~xgwang/CUHK_identification.html}} \cite{Li-ACCV-2012}
was captured with two camera views in a campus environment. Different from the above datasets, images in this dataset are of higher resolution. The CUHK Campus dataset contains 971 persons, and each person has
two images in each camera view. Camera A captures the frontal view or back view of pedestrians, while camera B captures the side views. All images were scaled to $160\times60$ pixels. The persons were split to 485 for training and 486 for test (multi-shot). The results are shown in Fig. \ref{fig:cuhk01}. Our method largely outperforms existing state of the art methods. The best rank-1 identification rate reported to date is 34.30\% \cite{Zhao-CVPR-2014-MidLF}, while we has achieved 63.21\%, with an improvement of 28.91\%.

\subsection{Experiments on CUHK03}
The CUHK03 dataset \cite{Li-CVPR-2014-DeepReID} includes
13,164 images of 1,360 pedestrians. It is currently the largest publicly available person re-identification dataset. The CUHK03 dataset was captured with six surveillance cameras over months, with each person observed by two disjoint camera views and having an average of 4.8 images in each view. In addition to manually cropped pedestrian images, samples detected with a state-of-the-art pedestrian detector is also provided. This is a more realistic setting considering misalignment, occlusions and body part missing.% Illumination changes were caused by weather, sun directions, and shadow distributions even within a single camera view.

We run our algorithm with the same setting of \cite{Li-CVPR-2014-DeepReID}. That is, the dataset was partitioned into a training set of 1,160 persons and a test set of 100 persons. The experiments were conducted with 20 random splits and all the CMC curves were computed with the single-shot setting. The rank-1 identification rates of various algorithms in both labeled and detected setting are shown in Table \ref{tab:cuhk03}. The proposed method achieved 52.20\% and 46.25\% rank-1 identification rates with the labeled bounding boxes and the automatically detected bounding boxes, respectively, which clearly outperform the state-of-the-art method DeepReID \cite{Li-CVPR-2014-DeepReID}, with an improvement of 31.55\% for the labelled setting, and 26.36\% for the detected setting.
\begin{table}
    \centering
    \caption{Comparison of state-of-the-art rank-1 identification rates (\%) on the CUHK03 database \cite{Li-CVPR-2014-DeepReID} with both labeled and detected setting (P=100). The compared results are from \cite{Li-CVPR-2014-DeepReID}.}\label{tab:cuhk03}
\begin{tabular}{|c|c|c|}
  \hline
  % after \\: \hline or \cline{col1-col2} \cline{col3-col4} ...
   & \textbf{Labeled} & \textbf{Detected} \\
  \hline
  LOMO+XQDA & \textcolor{red}{\textbf{52.20}} & \textcolor{red}{\textbf{46.25}} \\
  DeepReID \cite{Li-CVPR-2014-DeepReID} & \textcolor{blue}{20.65} & \textcolor{blue}{19.89} \\
  KISSME \cite{kostinger2012large} & 14.17 & 11.70 \\
  LDML \cite{Guillaumin-ICCV-09} & 13.51 & 10.92 \\
  eSDC \cite{zhao2013unsupervised} & 8.76 & 7.68 \\
  LMNN \cite{weinberger2006distance} & 7.29 & 6.25 \\
  ITML \cite{davis2007information} & 5.53 & 5.14 \\
  SDALF \cite{Bazzani-CVIU-2013-SDALF} & 5.60 & 4.87 \\
  \hline
\end{tabular}
\end{table}

%In order to compare the learning capacity and generalization capability of different learning methods, we did another experiment by adding 933 images of 107 pedestrians to the training set, while keep the test set unchanged. Therefore, the training set has 1, 267 persons. These additional 933 images are captured from four camera views different from those in the test set. Adding training samples, which do not accurately match the photometric and geometric transforms in the test set, makes the learning more difficult.

\subsection{Analysis of the Proposed Method}
To better understand the proposed method, we analyze it in several aspects: role of Retinex, role of the local maximal occurrence operation, influence of subspace dimensions, and the running time. The analysis was performed on the VIPeR database, by randomly sampling a training set of 316 persons, and a test set of the remaining persons.

\subsubsection{Role of Retinex}
\begin{figure}
\centering
\includegraphics[width=26mm]{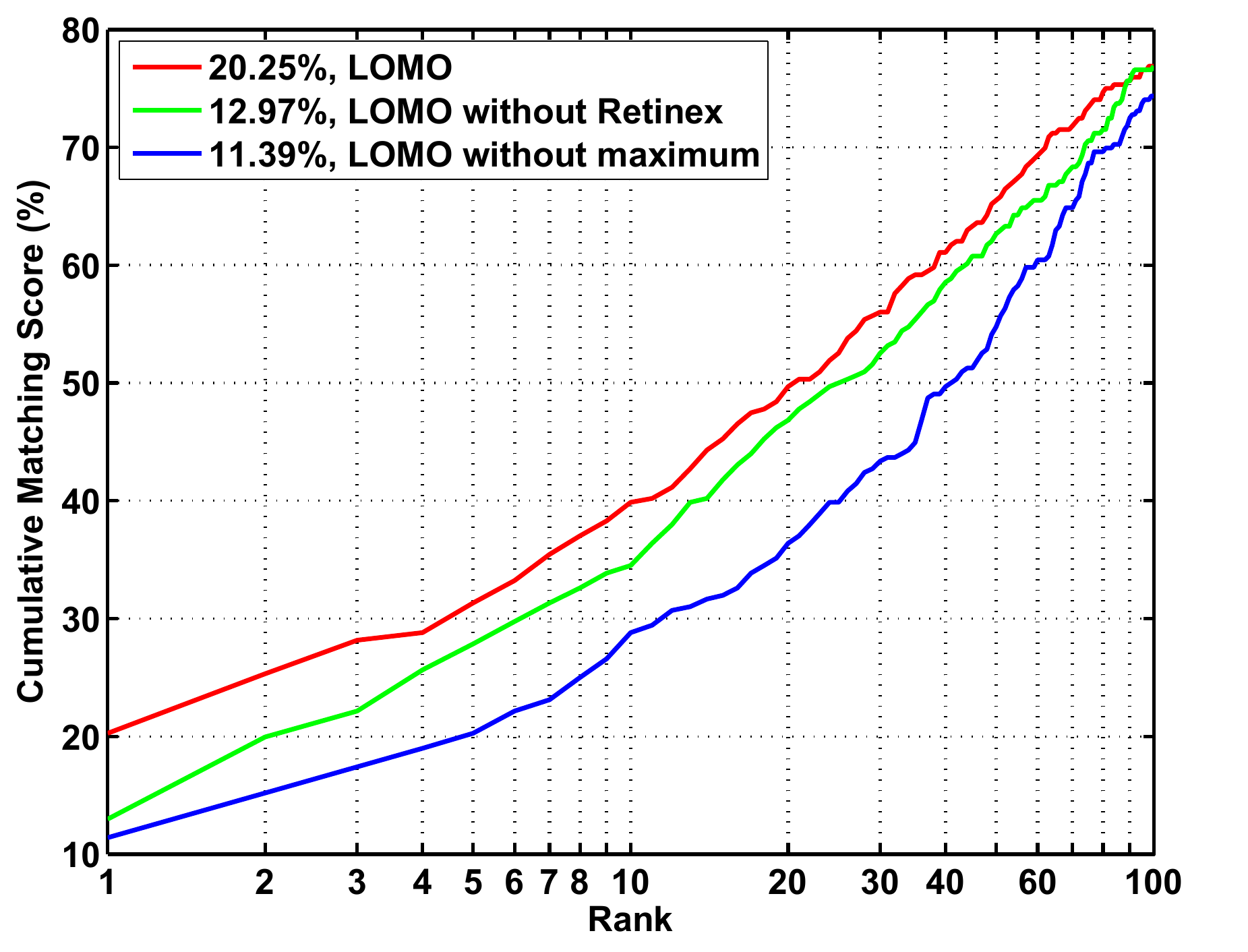}
\includegraphics[width=26mm]{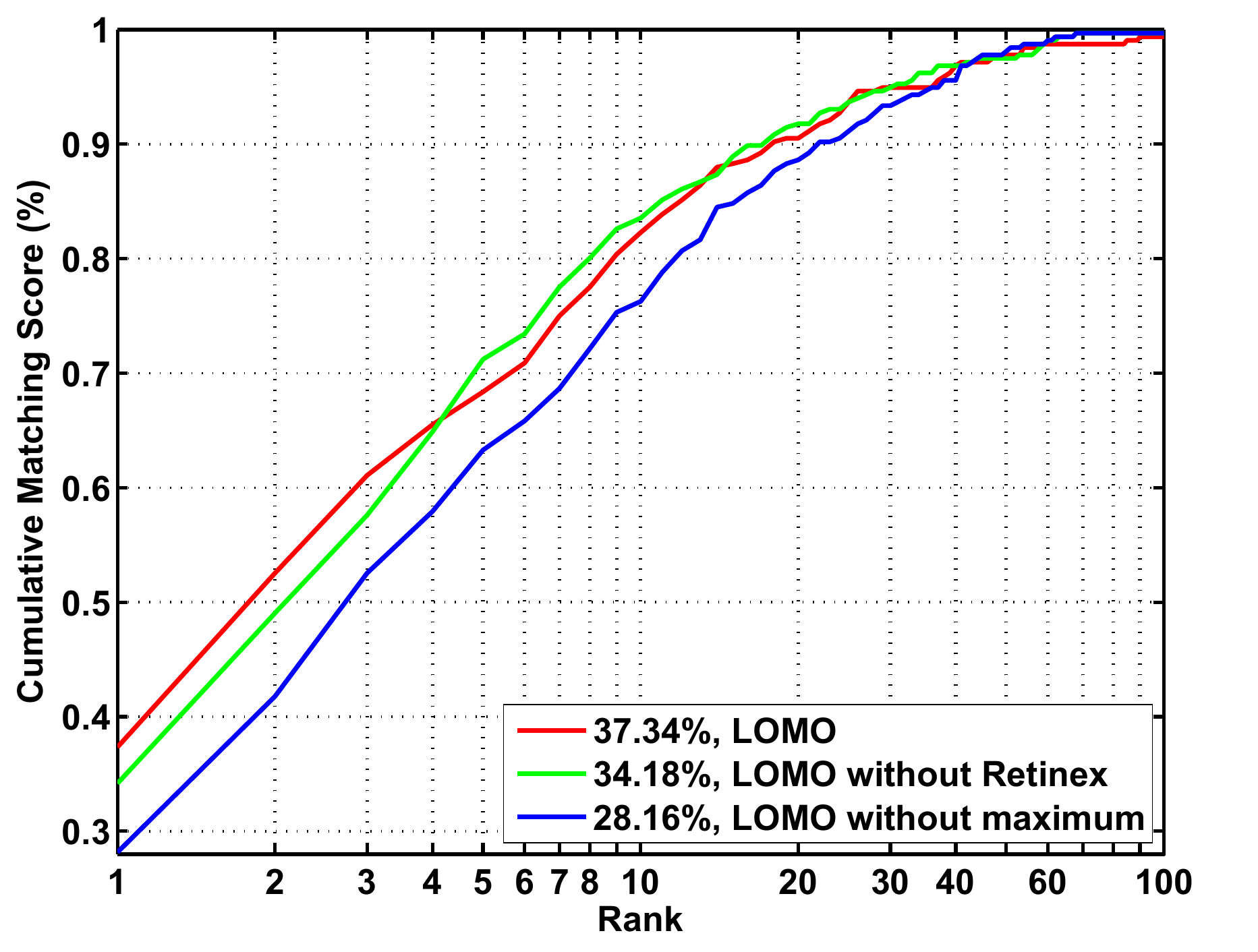}
\includegraphics[width=26mm]{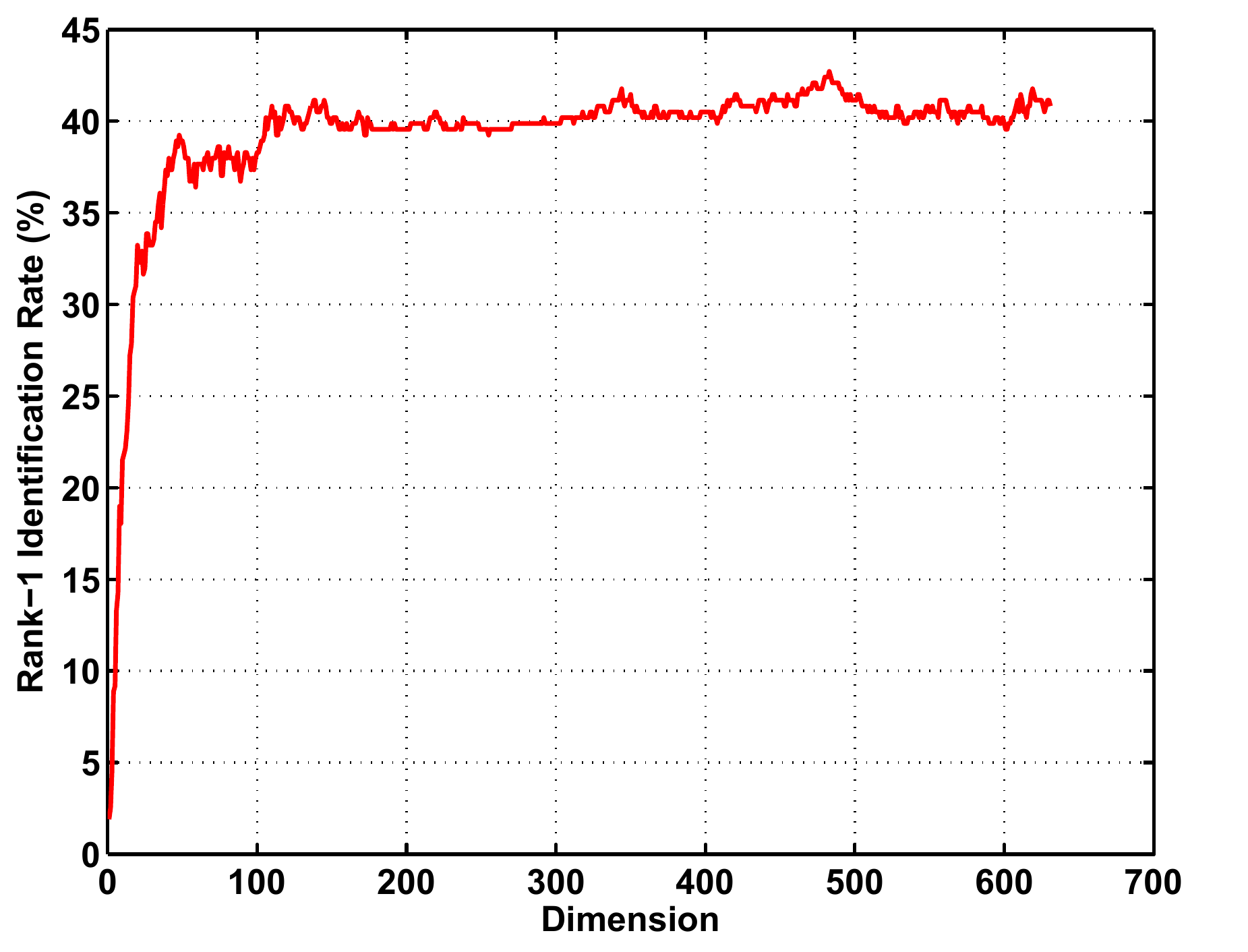}\\
(a) \hspace{21mm} (b) \hspace{21mm} (c)
\caption{CMC curves comparing the proposed feature with and without Retinex and the local maximal occurrence operation ((a) Cosine and (b) XQDA). (c) Rank-1 accuracy with varying subspace dimensions for the XQDA algorithm with LOMO feature.}\label{fig:role-retinex}
\end{figure}

We compared the proposed LOMO feature with and without the Retinex preprocessing, with results shown in Fig.~\ref{fig:role-retinex} (a) and (b). This comparison was done by using the direct Cosine similarity measure and the XQDA algorithm, respectively. From Fig.~\ref{fig:role-retinex} (a) we can see that, for direct matching, the performance can be obviously improved by applying the Retinex transform, with rank-1 accuracy being 12.97\% without Retinex, and 20.25\% with Retinex. This result indicates that Retinex helps to derive a consistent color representation across different camera views, as can also be observed from Fig.~\ref{fig:viper} (b). However, From Fig. \ref{fig:role-retinex} (b) it can be seen that the two features are boosted by XQDA to a similar performance. This may indicate that XQDA is able to learn a robust metric against illumination variations.

\subsubsection{Role of Local Maximal Occurrence}
The person re-identification performance is largely affected by viewpoint changes, which should be addressed in feature design or classifier learning. The proposed local maximal occurrence feature extraction is a strategy towards pose or viewpoint robust feature representation. By comparing the proposed feature with and without the local maximal occurrence operation, we find that this operation does largely improve the performance of cross-view person re-identification, as shown in Fig.~\ref{fig:role-retinex} (a) and (b). Without the local maximal occurrence operation, the rank-1 accuracy by applying the Cosine similarity measure (Fig.~\ref{fig:role-retinex} (a)) is 11.39\%, while applying this strategy, the rank-1 accuracy is improved to 20.25\%. When further applying XQDA (Fig.~\ref{fig:role-retinex} (b)), the performance gap is reduced, but the proposed feature still performs quite better with the local maximal occurrence operation than without it.

\subsubsection{Subspace Dimensions}
For the proposed XQDA algorithm, the dimension of the learned subspace has an influence in performance. This influence is shown in Fig.~\ref{fig:role-retinex} (c), obtained by applying XQDA with different subspace dimensions on the VIPeR dataset. Roughly, the performance is increasing with increasing dimensions, but it becomes stable after 100 dimensions. Therefore, it is not too difficult to determine a proper number of subspace dimensionality. We use an automatic way as specified by accepting all eigenvalues of $\Sigma_I^{-1} \Sigma_E$ that are larger than 1, which works quite well in all the experiments. However, one can also select a small value considering the computational complexity. As can be observed from Fig.~\ref{fig:role-retinex} (c), the rank-1 accuracy is consistently larger than 30\% when the subspace dimensions are larger than 16.
%
%\begin{figure}
%\centering
%\includegraphics[width=60mm]{xqda-dims}
%\caption{Rank-1 accuracy as a function of the subspace dimension for the proposed XQDA algorithm.}\label{fig:xqda-dims}
%\end{figure}
%

\subsubsection{Running Time}\label{subsubsec:time}
The training time comparison of metric learning algorithms is shown in Table~\ref{tab:time} (including subspace learning time). The training time was averaged over 10 random trials on the VIPeR dataset. All algorithms are implemented in MATLAB. The LMNN algorithm has MEX functions implemented in C or C++ to accelerate the computation. The training was performed on a desktop PC with an Intel i5-2400 @3.10GHz CPU. Table \ref{tab:time} shows that the KISSME, RLDA, and XQDA algorithms, which have closed-form solutions, are very efficient, while ITML and LMNN, which require iterative optimizations, are time consuming.
\begin{table}
\centering
\caption{Training time (seconds) of metric learning algorithms.}\label{tab:time}
\begin{tabular}{|c|c|c|c|c|c|}
  \hline
  % after \\: \hline or \cline{col1-col2} \cline{col3-col4} ...
   & XQDA & KISSME & RLDA & ITML & LMNN \\
  \hline
  Time & 1.86 & 1.34 & 1.53 & 36.78 & 265.28 \\
  \hline
\end{tabular}
\end{table}

Besides, we also evaluated the running time of the proposed feature extractor. In processing $128\times48$ person images, the LOMO feature extractor requires 0.012 seconds per image on average, which is very efficient. This code is also implemented in MATLAB, with a MEX function implemented for Retinex. Considering the effectiveness and efficiency of both the proposed LOMO feature and XQDA algorithm, we release both codes\footnote{\url{http://www.cbsr.ia.ac.cn/users/scliao/projects/lomo_xqda/}} for future research and benchmark on person re-identification.

\section{Summary and Future Work}\label{sec:summary}
In this paper, we have presented an efficient and effective method for person re-identification. We have proposed an efficient descriptor called LOMO, which is shown to be robust against viewpoint changes and illumination variations. We have also proposed a subspace and metric learning approach called XQDA, which is formulated as a Generalized Rayleigh Quotient, and a closed-form solution can be obtained by the generalized eigenvalue decomposition. Practical computation issues for XQDA have been discussed, including the simplified computation, the regularization, and the dimension selection. Experiments on four challenging person re-identification databases, VIPeR, QMUL GRID, CUHK Campus, and CUHK03, show that the proposed method improves the state-of-the-art rank-1 identification rates by 2.2\%, 4.88\%, 28.91\%, and 31.55\% on the four databases, respectively. Due to the promising performance of the LOMO feature, it would be interesting to study other local features (e.g. Gabor, other color descriptors, etc.) or feature coding approaches with the same LOMO idea for person re-identification. It is also interesting to see the application of XQDA to other cross-view matching problems, such as the heterogeneous face recognition.

\section*{Acknowledgments}
This work was supported by the Chinese National Natural Science Foundation Projects \#61203267, \#61375037, \#61473291, National Science and Technology Support Program \#2013BAK02B01, Chinese Academy of Sciences Project No. KGZD-EW-102-2, and AuthenMetric R\&D Funds.

{\small
\bibliographystyle{ieee}
\bibliography{../../Bib/Liao}
}

\end{document}